\documentclass[conference,a4paper]{IEEEtran}
\IEEEoverridecommandlockouts

\usepackage[hidelinks]{hyperref}
\usepackage[cmex10]{amsmath}
\usepackage{amssymb,amssymb,amsfonts}
\usepackage{dblfloatfix}

\usepackage[ruled,vlined]{algorithm2e}
\usepackage{graphicx}
\graphicspath{{Figures/PDF/}{Figures/PNG/}}

\usepackage{booktabs}
\usepackage{siunitx}
\usepackage[numbers,compress]{natbib}
\usepackage{texnames}
\usepackage{bm,bbm}
\usepackage{orcidlink}
\usepackage{algorithmic}
\usepackage{textcomp}
\usepackage{xcolor}
\usepackage{subcaption}
\usepackage{cleveref}
\usepackage{adjustbox}
\usepackage{multirow}
\usepackage{dsfont}
\usepackage{float}
\newcommand{\etal}{\textit{et al}.}

\begin{document}
\title{\uppercase{Surfel-based 3D Registration with Equivariant SE(3) Features}
\thanks{Thanks for Graduate Research Scholarships of the University of Melbourne to sponsor Xueyang for his PhD study at Melbourne.}
}

\author{	\IEEEauthorblockN{Xueyang Kang \orcidlink{0000-0001-7159-676X
}}
	\IEEEauthorblockA{\textit{University of Melbourne, KU Leuven}\\
		Parkville VIC 3052, Austraslia \\ 
		alex.kang@kuleuven.be}
	\and
	\IEEEauthorblockN{Hang Zhao \orcidlink{https://orcid.org/0000-0002-5279-0273
}}
	\IEEEauthorblockA{\textit{University of Melbourne}\\
		Parkville VIC 3052, Austraslia\\
		h.zhao438331981@gmail.com}
	\and
	\IEEEauthorblockN{Kourosh Khoshelham\orcidlink{0000-0001-6639-1727}}
	\IEEEauthorblockA{\textit{University of Melbourne}\\
		Parkville VIC 3052, Austraslia\\
		k.khoshelham@unimelb.au}
	\and
	\IEEEauthorblockN{Patrick Vandewalle\orcidlink{0000-0002-7106-8024}}
	\IEEEauthorblockA{\textit{KU Leuven}\\
		3000 Leuven, Belgium\\
		p.vandewalle@kuleuven.be}
}
\vspace{-1.6em}

\maketitle
\begin{abstract}
Point cloud registration is crucial for ensuring 3D alignment consistency of multiple local point clouds in 3D reconstruction for remote sensing or digital heritage. While various point cloud-based registration methods exist, both non-learning and learning-based, they ignore point orientations and point uncertainties, making the model susceptible to noisy input and aggressive rotations of the input point cloud like orthogonal transformation; thus, it necessitates extensive training point clouds with transformation augmentations. To address these issues, we propose a novel surfel-based pose learning regression approach. Our method can initialize surfels from Lidar point cloud using virtual perspective camera parameters, and learns explicit $\mathbf{SE(3)}$ equivariant features, including both position and rotation through $\mathbf{SE(3)}$ equivariant convolutional kernels to predict relative transformation between source and target scans. The model comprises an equivariant convolutional encoder, a cross-attention mechanism for similarity computation, a fully-connected decoder, and a non-linear Huber loss. Experimental results on indoor and outdoor datasets demonstrate our model superiority and robust performance on real point-cloud scans compared to state-of-the-art methods.
\end{abstract}


\section{Introduction}
Point cloud registration is crucial in 3D reconstruction, shape pose estimation, digital twin in remote sensing, and various applications, aiming to estimate relative transformations between source and target 3D point cloud scan pairs. These point clouds are usually generated from 3D scan sensor like LiDAR, structured light, RGB-D or stereo camera.\cite{izadi2011kinectfusion, huai2015real}. 

Registration algorithms are generally categorized as rigid or non-rigid. Rigid methods, such as Iterative Closest Point (ICP) and its variants \cite{segal2009generalized}, estimate transformations through point-wise error optimization. Non-rigid approaches consider potential deformations or non-linear distortions, catering to scenarios with deformable objects or dynamic environments \cite{stuckler2014multi, myronenko2010point}.
Traditional methods rely on non-learning-based iterative optimizations such as ICP \cite{low2004linear}, which are often slow to converge, and suffer from high outlier-inlier ratios. In contrast, deep learning-based techniques convert raw input points into high-dimensional feature spaces for direct correspondence establishments, as demonstrated by DGR \cite{choy2020deep} and PointDSC \cite{bai2021pointdsc}. SpinNet \cite{ao2021spinnet} and RoReg \cite{wang2023roreg} are either $\mathbf{SO(2)}$ or $\mathbf{SO(3)}$ equivariant, to learn rotation and position from point cloud jointly. But they may fail to predict registration pose given noisy points, or challenging sensor motion, \emph{e.g.}, in-image-plane rotation or camera motion parallel to image plane.

To address these challenges, we employ surfel representation, the small, oriented disks un-projected from depth map aligned with view image, or from LiDAR point cloud. 
Dahl \etal \cite{surfel_geometry} first demonstrates use of surfel in large-scale 3D reconstruction, Behley and Stachniss \cite{behley2018efficient} apply it to self-driving scenarios. Pfister \emph{et al.} \cite{surfel_primitives} further leverage surfels for high-fidelity mesh surfaces in 3D graphical rendering. Surfels can be considered as 2D Gaussians. Compared to point clouds, the 2D surfel-based Gaussian approach exhibits superior robustness by leveraging data uncertainties, as evidenced by the recent Gaussian Splatting technique \cite{kerbl20233d}. Furthermore, we introduce a specialized equivariant network model to learn $\mathbf{SE(3)}$ equivariant features, including rotation and translation from surfel input through E2PN \cite{chen2021equivariant}, Pairwise equivariant features undergo cross attention to create an attention-based similarity feature map for correspondence establishment. These features are then processed through fully-connected layers to estimate relative transformation for alignment. Lastly, we also introduce a specific $\mathbf{SE(3)}$ differentiable Huber loss function for surfel-based registration. 
\begin{figure*}[!th]
\vspace{-1.6em}
\centering 
\includegraphics[width=0.92\linewidth]{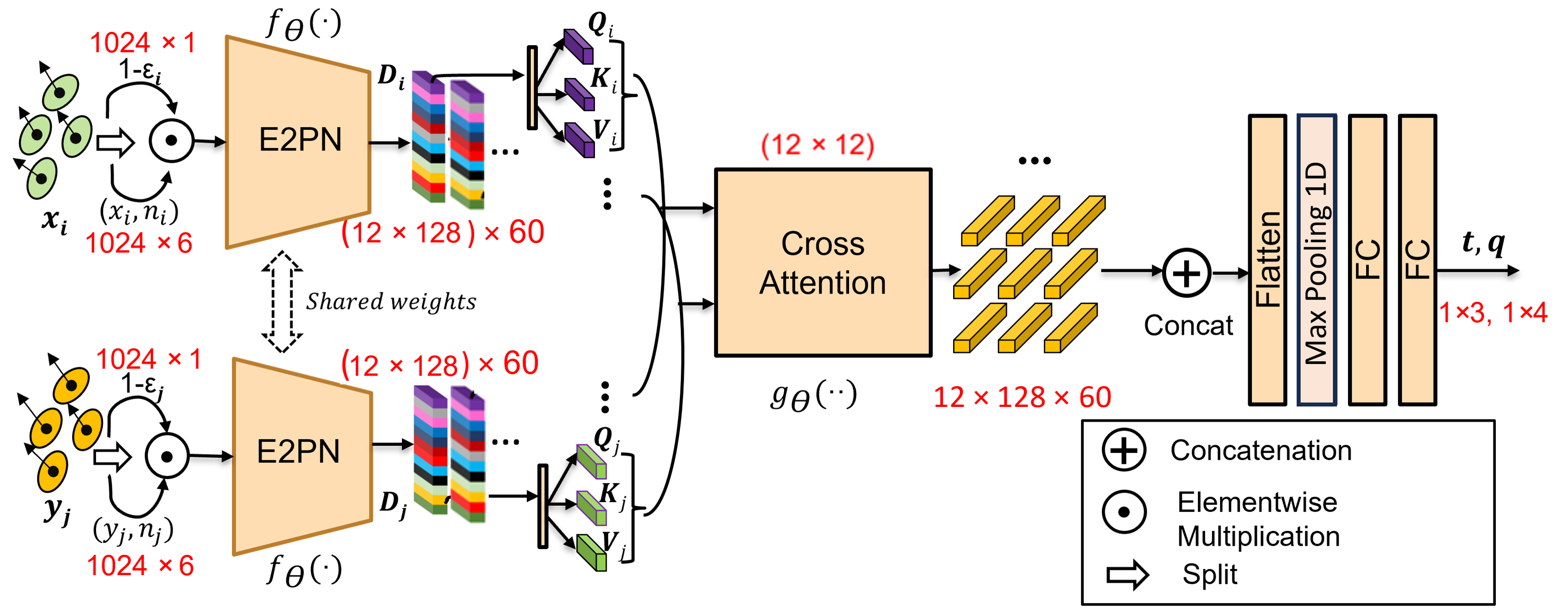}
\caption{The network structure features a shared encoder for surfels (6 dimensions plus uncertainty radius) from both source and target frames in $\mathbf{SE(3)}$ space. This encoder maps 1024 surfels with 6 dimensions (position and normal), weighted by confidence value $(1-\epsilon(\cdot))$, into 12 feature descriptors in 128-dim with 60 group rotation orders. Then each descriptor undergoes linear embedding to produce triplet token embeddings $\mathbf{Q}$, $\mathbf{K}$, and $\mathbf{V}$. Cross-attention $g_{\theta}(\cdot)$ is applied to feature descriptors from source and target frames in 12-channel dimension. The resulting tokens are in shape of (12 $\times$ 128 $\times$ 60), where each token in 128-dim $\times$ 60 order groups, and the attention map is 12 $\times$ 12, where each element token in the attention map is formulated via descriptor dot product. Then the features are flattened and processed through Fully-Connected (FC) layers, mapping features to relative position $\mathbf{t}{}^{}$ and relative quaternion $\mathbf{q}{}^{}$ rotation. A close-up of E2PN module is provided below.}
\label{fig:network-structure}
\vspace{-1.5em}
\end{figure*}

\section{Related Work}
Despite significant progress in point cloud representations for 3D reconstruction \cite{kang20183d}, existing methods often lack built-in equivariance, limiting their robustness in 3D registration tasks. While prior works on 3D registration have explored both traditional optimization-based approaches and deep feature learning, they do not explicitly incorporate rotational equivariance at the architectural level. 
\noindent$\textbf{\emph{Applications of 3D Surfels}}$ SurfelMeshing \cite{schops2019surfelmeshing} has been applied to 3D mapping in indoor scenes, while other approaches have used surfels to large-scale outdoor mapping \cite{yuan2022efficient}. 

\noindent$\textbf{\emph{3D Registration}}$ Registration techniques are widely used in shape alignment \cite{choy2020deep, zhou2016fast}, remote sensing \cite{yan2024amalgamating, ye2022multiscale} and deformable target scanning \cite{bhatnagar2020loopreg}. Traditional non-learning methods like Iterative Closest Point (ICP), kiss-ICP \cite{low2004linear}, and point-to-plane ICP \cite{park2003accurate} struggled with nonlinear optimization errors. In contrast, state-of-the-art deep learning models like PointDSC \cite{bai2021pointdsc} and Deep Global Registration (DGR) \cite{choy2020deep} explore searching correspondences in high-dimensional feature spaces or Superpoint descriptors \cite{li2024improved}. Max-Clique \cite{zhang20233d}, GeoTransformer \cite{qin2023geotransformer}, and Equi-GSPR \cite{kang2025equi} leverage the latest graph or attention learning to create network backbone. 

\noindent$\textbf{\emph{Equivariant Feature Representation}}$ Equivariant models have emerged as robust solutions for 3D applications such as point cloud registration and pose estimation. These models maintain $\mathbf{SE(3)}$ rotational and translational equivariance, essential for consistent alignment performance. Examples include Spherical CNNs \cite{cohen2018spherical}, group CNNs \cite{finzi2020generalizing}, and SE(3)-Transformers \cite{fuchs2020se}.

SpinNet \cite{ao2021spinnet} and RoReg \cite{wang2023roreg} leverage rotation-equivariant features for point cloud registration. 
\section{Method}
For perspective image input, we can initialize surfels from the unprojected depth map and its associated normal map. For Lidar, we create a virtual camera plane in front of the sensor with fixed Field of View (FoV), then the surfel uncertainties can be calculated based on a sensor virtual perspective projection model, accounting for both inverse distance uncertainty and Lidar ray direction angle to the sensor principal axis. The whole initialization process serves as pre-processing. For LiDAR point cloud, surfels can be created from the neighboring non-coplanar triplet points for normal vector estimation, and 1D uncertainty can be derived proportionally from the point density.  To ensure registration efficiency, surfels from source and target frames are downsampled before feeding into the neural network model. The model architecture, as illustrated in \cref{fig:network-structure}, consists of three components: an $\mathbf{SE(3)}$ equivariant convolution kernel-based encoder, a cross-attention, and a decoder for predicting relative transformation.

\subsection{Surfel Initialization}
Each surfel in the source frame, indexed by $i$, consists of three main components: the 3D position $\mathbf{x}_i \in \mathbb{R}^3$, the normal vector $\mathbf{n}_i \in \mathbb{R}^3$, and a scalar radius $\epsilon_i \in \mathbb{R}$. The surfel $\mathbf{y}_j$ is in the target frame. The surfel center position is determined by the point cloud position. To calculate the normal map from the point cloud, a parallel CUDA-based PCA plane normal initialization from the K-nearest-neighboring points is employed. The normal determines the orientation of the surfel disk. The surfel radius $\epsilon_i$ is then derived by the expression:
\begin{equation}
   \epsilon_{i} = \gamma \cdot  C \frac{e^{-\hat{\rho}}}{1+e^{-\tan(\theta)}},   
\label{eq:surfel_radius}
\end{equation}
where $C$ is the normalization factor, $\gamma$ is the point intensity normalied to probability, and $\theta$ represents the view angle between the virtual camera principal axis $\vec{o}$ and the ray $\vec{r}$ emitted from the sensor center through the pixel location, as expressed in the following equation,
\begin{equation}
    \theta = \arccos\left(\frac{\vec{r} \cdot \vec{o}}{\lVert \vec{r} \rVert \lVert \vec{o} \rVert}\right).
    \label{eq:angle}
\end{equation}

\begin{figure}[!ht]   
\vspace{-0.5em}
\begin{subfigure}{0.31\linewidth}
  \includegraphics[width=\linewidth]{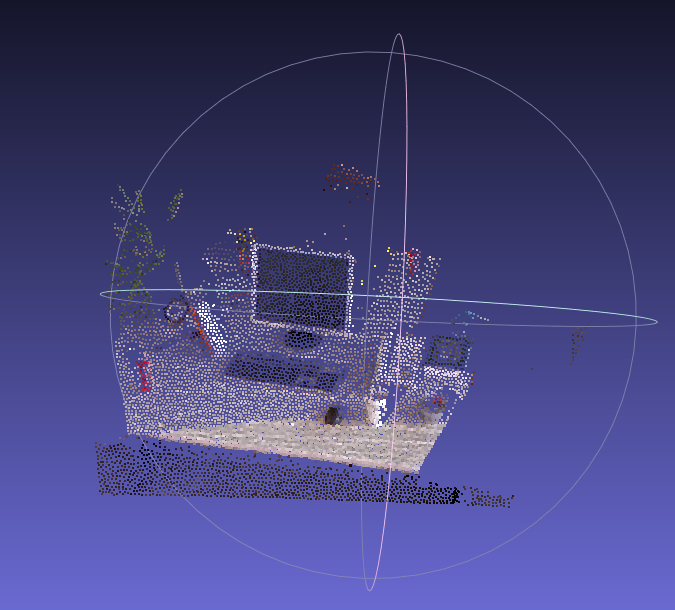}
    \caption{Colored 3D points}
    \vspace{-0.0em}
    \label{fig:depth}
\end{subfigure}
\begin{subfigure}{0.31\linewidth}
    \includegraphics[width=\linewidth]{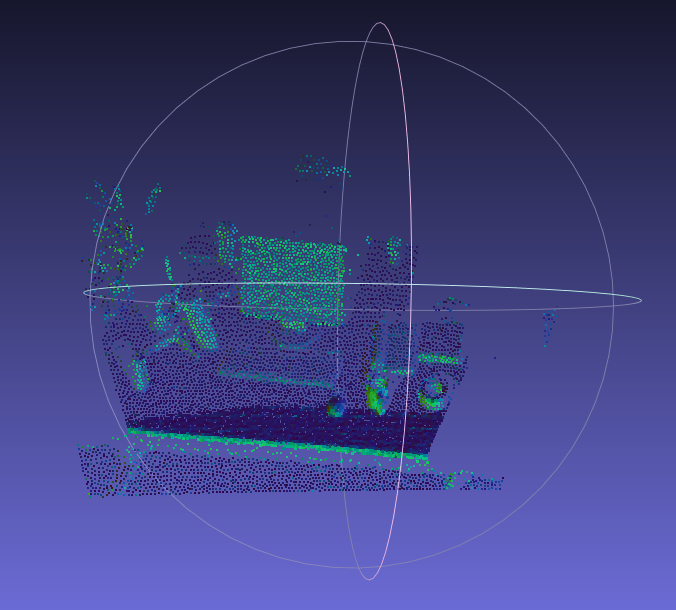}
    \caption{Normals of 3D points}
    \vspace{-0.0em}
    \label{fig:normal}
\end{subfigure}
\begin{subfigure}{0.31\linewidth}
    \includegraphics[width=\linewidth]{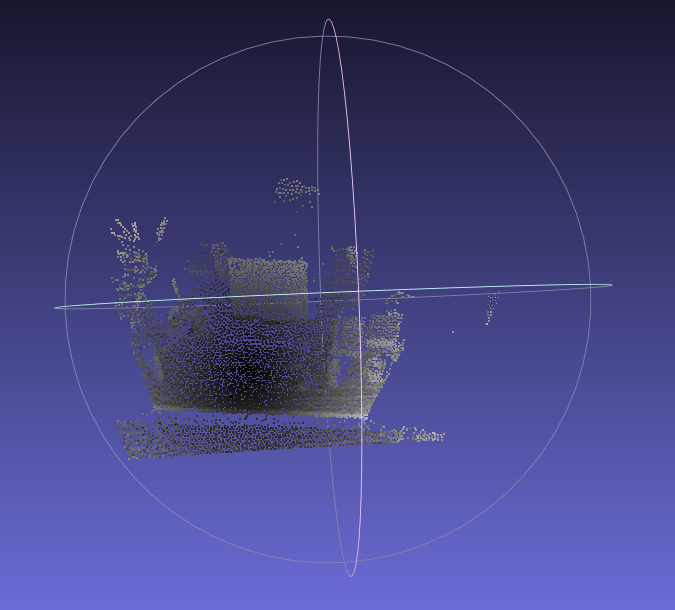}
    \caption{Surfel Uncertainties}
    \label{fig:surfels}
\end{subfigure}
\caption{Surfel component visualization, from left to right is surfel point position, surfel normals and uncertainties, where each point position corresponds to a disk center, with the normal at that position determining surfel orientation, and the surfel radius represents the uncertainty (black indicate small uncertainties, while white depicts the large uncertainties).} 
\label{fig:dsurfels}
\vspace{-1.2em}
\end{figure}

The value of $\rho$ represents the inverse of the depth, which is then truncated to $\hat{\rho}$ within the inverse depth range $(\rho_{min},\rho_{max})$ of the sensor,
\begin{equation}
    \hat{\rho} = \min\left(\max\left(\rho, \rho_{min}^{}\right), \rho_{max}\right).
\end{equation}
The colored point cloud position, the normals calculated from the neighboring points of interest points, and the initialized surfel uncertainties (radius) overlaid onto points are presented separately in \cref{fig:dsurfels}. This surfel radius heuristic initializes uncertainty with the virtual sensor perspective model, resulting in smaller radii for points near the image centre and larger radii for points further away. Additionally, the radius increases with depth distance, modelling depth measurement noise.

\subsection{Network Structure}
Given initialized surfels of source frame $\mathbf{s}_{i}^{} \in {\mathbf{s}_{1}^{},...,\mathbf{s}_{N}^{}}$ and target frame $\mathbf{s}_{j}^{} \in {\mathbf{s}_{1}^{},...,\mathbf{s}_{N}^{}}$, all surfels are encoded by the same encoder $f_{\theta_{}}(\cdot)$. Notably, the position and normal vectors $\mathbf{n}_{(\cdot)}$, $\mathbf{p}_{(\cdot)}$ of each surfel are weighted by a factor of $(1-\| \epsilon(\cdot) \|)$ to reduce the influence of highly uncertain surfels. The encoder architecture is augmented on E2PN \cite{chen2021equivariant}, with doubled feature dimensions compared to the original point cloud input. Equivariance is maintained through a symmetric conv-kernel $\mathcal{\kappa}$ arranged in an icosahedral shape solids. \(\mathbf{SE}(3)\) is defined as Special Euclidean Group, the group of rigid body transformations in 3D space, including \textit{rotations} and \textit{translations}. An element of \(\mathbf{SE}(3)\) is represented as a matrix \(3 \times 4\), $\mathbf{T} = [\mathbf{R}, \mathbf{t}]$,
where \(\mathbf{R} \in \mathbf{SO}(3)\) is a \(3 \times 3\) rotation matrix, and \(\mathbf{t} \in \mathbb{R}^3\) is a translation vector. \(\mathbf{SO}(3)\): Special Orthogonal Group in 3D is the group of all 3D rotations. 
In the following convention, the symbol $'$ next to a symbol indicates discretization operation. E2PN encoder features are aligned in the spherical space $\mathbf{S}^{2'}\times\mathcal{R}^3$, where coordinates of each feature vertex in $\mathcal{R}^3$, associated with the 128-dimension feature descriptor, and $\mathbf{S}^{2'}$ signifies the discretized sphere surface. This discretized feature representation is determined by $\mathbf{SO}(3)'/\mathbf{SO}(2)'$, where $\mathbf{SO}(2)'$ is a subgroup of $\mathbf{SO}(3)$. The quotient space is defined as a group of rotations $\mathbf{R}_{i/j}$ with a same endpoint, such as the sphere north pole after rotation. This discretization of $\mathbf{SO}(3)$ facilitates more efficient and accelerated learning. The $\mathbf{SE}(3)$ feature is constructed by extending the $\mathbf{SO}(3)$ rotation feature, incorporating translation through the concatenation of point coordinates. \\
\begin{figure}[!ht]   
\vspace{-2.2em}
\centering
  \includegraphics[width=0.86\columnwidth]{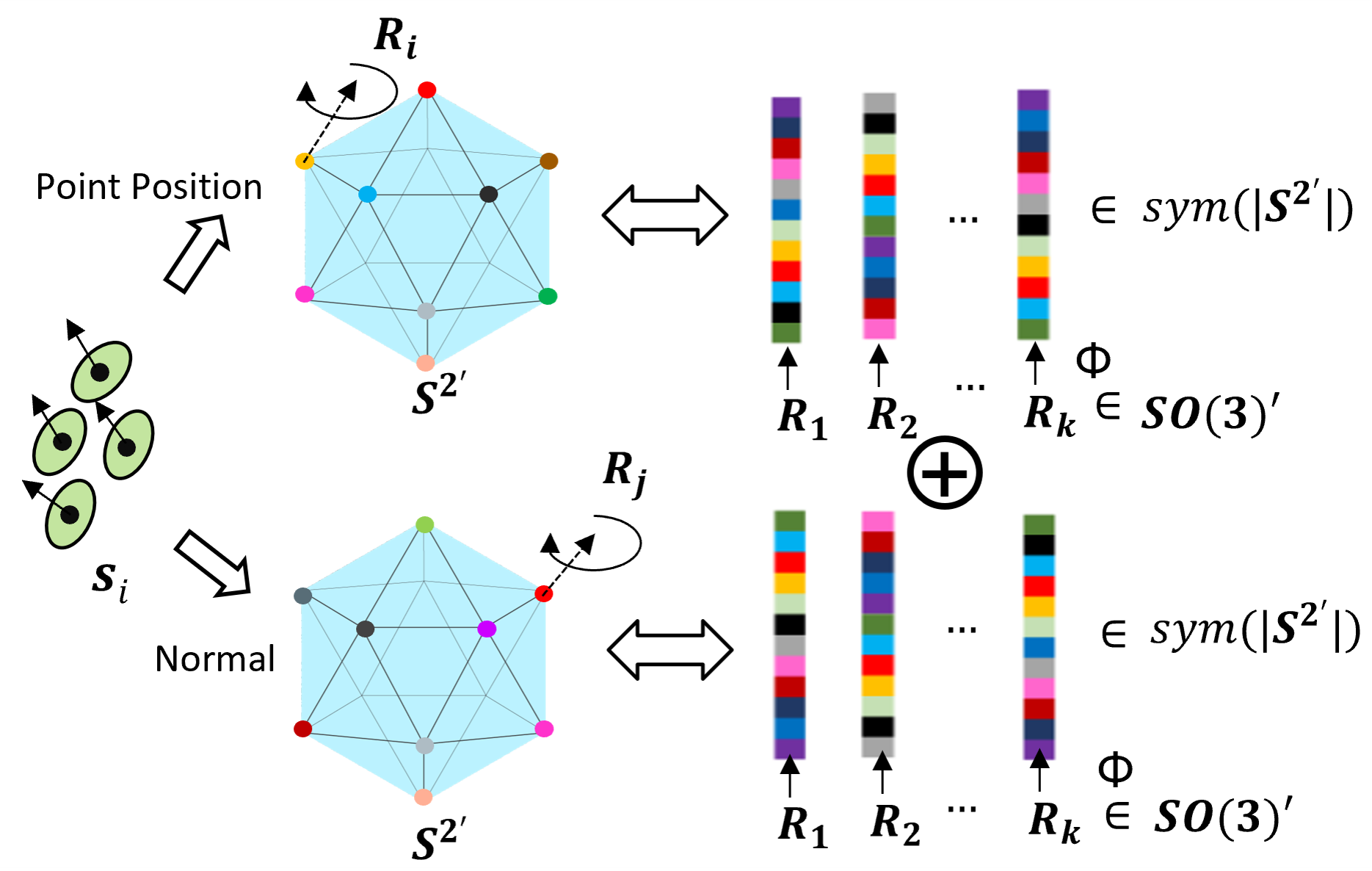}
    \caption{E2PN encoder learns equivariant features from the surfel input. The learning of equivariant features is implemented in two branches for learning surfel position and surfel normals separately. The discretized $\mathbf{SO(3)'}$ can be recovered from the quotient feature $\mathbf{S}^{2'}$ by permutation order of 12 vertices (here features are simplified into 1D dimension of rotation order for demo) on Platonic solids, as an approximation of the group rotation in 60 orders.}
    \vspace{-0.0em}
    \label{fig:kernel}
\vspace{-1.0em}
\end{figure}
The surfel undergoes convolution with two distinct symmetric kernels, $\kappa_1$ and $\kappa_2$, as shown in \cref{fig:kernel}, relating to point position and normal respectively. These are concatenated post-convolution. The icosahedron comprises 60 rotations, each denoted by various permutation orders $\mathbf{R}_{(\cdot)}$. The E2PN symmetric kernel selects one rotation from the 60 options to generate the output equivariant feature. This is achieved by choosing the maximum sum of each rotation feature along the 12-channel dimension.

The transformation applied to the input  point cloud is denoted by group rotation \( g'(\cdot) \). The key idea of equivariant feature learning is that the output features of the encoder are transformed accordingly to preserve $\mathbf{SE(3)}$ equivariance $f_{\theta}(g'(x)) = g'(f_{\theta}(x))$. This ensures that the features are equivariant to the input transformations. 
The equ-features after the E2PN encoder of the source and target frames are denoted as $\mathbf{D}_i, i \in ({1,...,12})$ and $\mathbf{D}_j, j \in ({1,...,12})$, where each has 128 dimensions. Next, we use the linear layer to project each descriptor into a triplet composed of $\mathbf{Q}_{(\cdot)}, \mathbf{K}_{(\cdot)}, \mathbf{V}_{(\cdot)}$ tokens. We use the same index convention throughout the paper, where $i$ denotes the index of the source frame, and $j$ refers to the target frame. The triplet tokens, derived from feature descriptors at the 12 corners of icosahedral planoids (see \cref{fig:kernel}), are aggregated from neighboring surfel coordinates. This effectively fuses the input surfels into 12 distinct regions, which are then used in the subsequent cross-attention module. 
The cross-attention $g_{\theta}(\cdot)$ is then applied to calculate the attention-weighted equivariant features (attention map in $12 \times 12$) from the pairwise frames, finally to be decoded by the fully connected layers into the transformation estimation.

\subsection{Loss Function}
Inspired by the node-wise supervision in PointDSC \cite{bai2021pointdsc}, we adapt the original binary cross-entropy loss to a non-linear Huber loss. This maps the transformed point error into $\mathcal{L}_2$ norm when it is small and into $\mathcal{L}_1$ normal when the error is large. The point position from the source frame is transformed by the predicted rotation $\mathbf{R}$ and translation $\mathbf{t}$ into $\mathbf{y}_{j^*} = \mathbf{R}\hat{\mathbf{x}}_{i^*} + \mathbf{t}$. The rotation matrix $\mathbf{R}$ is derived from the predicted quaternion $\mathbf{q}$. The Huber loss is defined as below,
\begin{equation}
\label{eq:loss}
\mathcal{L}_\text{Huber} = \begin{cases}
\frac{1}{2}(\hat{\mathbf{x}}_{i^*}-\mathbf{y}_{j^*})^2, & \text{if } |\hat{\mathbf{x}}_{i^*}-\mathbf{y}_{j^*}| \leq \delta \\
\delta(|\hat{\mathbf{x}}_{i^*}-\mathbf{y}_{j^*}| - \frac{1}{2}\delta), & \text{otherwise}
\end{cases}
\end{equation}
The threshold is set to 0.6m. The correspondence index pair $(i, j)$ is established using the nearest neighboring point search. 
\begin{figure*}[!thbp]
    \centering
    \includegraphics[width=0.95\textwidth]{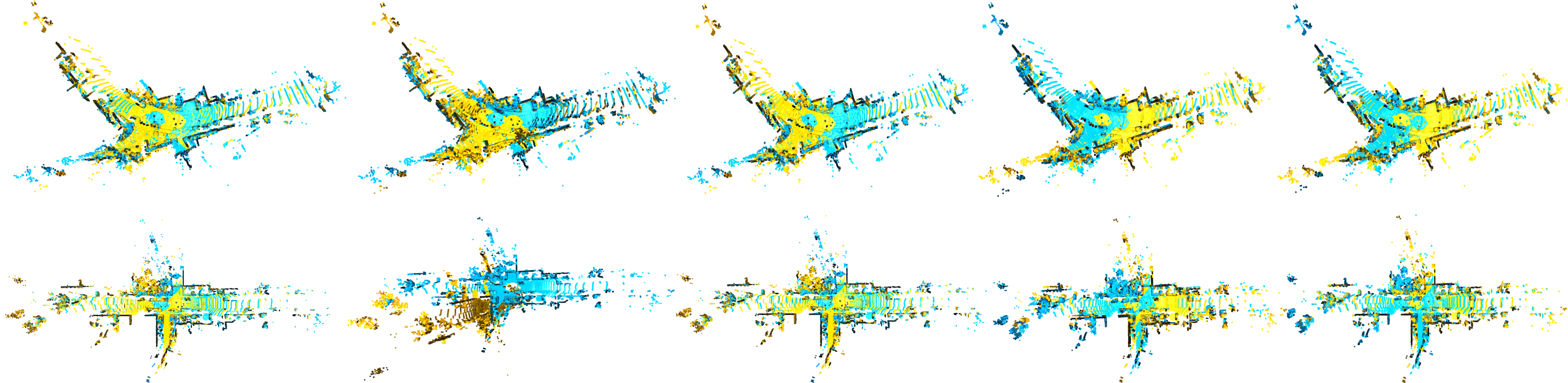}
        \hspace{6.4em} PointDSC \cite{bai2021pointdsc}
        \hspace{4.8em} MAC \cite{zhang20233d}
        \hspace{4.2em}  GeoTransformer \cite{qin2023geotransformer}
         \hspace{2.0em} Ours
         \hspace{4.6em} Ground Truth\\
    \caption{Comparison results on KITTI \cite{geiger2012we} . For each dataset, the top three models with good performance are presented visually.}
    \vspace{-1.6em} 
    \label{fig:cmp-3DMatch-KITTI}
\end{figure*}

\section{Experiment Results}
To evaluate the model performance, we utilized outdoor dataset KITTI \cite{geiger2012we}. Our model was trained on each dataset separately for fair comparison against other baseline models. 
We employed 3D voxel-based downsampling to generate 2048 points unprojected from the depth map of each frame for surfel initialization.
\begin{table}[!th]
\vspace{-1.0em}
\centering
    \caption{Baseline evaluation results on KITTI \cite{geiger2012we} under FPFH \cite{rusu2009fast} settings.}
    \label{tab:cmps-baselines}
\large
\vspace{-0.4em}
\begin{adjustbox}{width=\columnwidth}   
     \begin{tabular}{c ccc ccc}
        \toprule
        \multirow{2}{*}{Method} &  \multicolumn{6}{c}{KITTI \cite{geiger2012we}} \\
        \cmidrule(lr){2-7}
       & DGR \cite{choy2020deep} & SpinNet \cite{ao2021spinnet} & PointDSC \cite{bai2021pointdsc} & MAC \cite{zhang20233d} & GeoTransformer \cite{qin2023geotransformer} & Ours \\ 
        \midrule
       RE($^{\circ}$) $\downarrow$ & 1.67 & 0.98 & 0.35 & 0.40 & 0.27 & $\mathbf{0.25}$ \\ 
        \hline
       TE(cm) $\downarrow$ & 34.74 & 15.60 & 7.83 & 8.46 & 7.40 & $\mathbf{7.03}$ \\ 
        \hline
       RR(\%) $\uparrow$ & 73.69 & 81.44 & 90.27 & 99.46 &  99.80 & $\mathbf{99.85}$ \\ 
        \hline
        F1(\%) $\uparrow$ & 4.51 & 80.91 & 70.89  & 90.25 & 91.54 & $\mathbf{91.72}$ \\ 
        \bottomrule
    \end{tabular}
\end{adjustbox}    
\vspace{-0.4em}
\end{table}

\noindent\textbf{Evaluation Details.} We use the Rotation Error (RE) and Translation Error (TE) to evaluate the accuracy of rotation and translation separately. Furthermore, we incorporate the Registration Recall (RR) and \emph{F1 score} as registration success evaluation metrics. 
We compare our model with popular deep learning-based models, including PointDSC \cite{bai2021pointdsc}, Deep Global Registration \cite{choy2020deep} (DGR), Maximal Clique \cite{zhang20233d} (MAC). Additionally, we choose equivariant methods, like SpinNet \cite{ao2021spinnet}, RoReg \cite{wang2023roreg} and GeoTransformer \cite{qin2023geotransformer} for KITTI, but for RoReg, it does not provide pre-trained weights on KITTI, therefore, we trained it from scratch, but the results are really bad thus we are not reporting it in the paper. For all the feature point descriptor-based approaches, like DGR \cite{choy2020deep}, FPHF \cite{rusu2009fast} descriptor is used for evaluation on KITTI \cite{geiger2012we}. We provide quantitative evaluation results in \cref{tab:cmps-baselines}, and qualitative comparison results including top three baseline models in \cref{fig:cmp-3DMatch-KITTI}. All these results exhibit superior performance of our model over baselines. 
\noindent\textbf{Ablation Study.} 1) We perform eight different types of ablation tests to verify the contribution of each design to our model performance, as shown in \cref{tab:ablation-cmps}. Point cloud position as input fed into various SOTA point cloud encoders, or vanilla E2PN encoder cannot achieve performance on par with our surfel-based equi-model design. The uncertainties and Huber loss are all beneficial to improving the model performance. 2) We provide the robustness analysis of input scans perturbed by various levels of rotation and translation in \cref{tab:robustness-analysis} from small to big transform perturbation. 3) Finally, we provide model complexity comparisons for the top five baseline models, and our model in \cref{tab:model-complexity} as shown below to showcase the low latency and small model size complexity of our model compared to other baselines.
\begin{table}[ht!]
\centering
    \caption{Ablation study on KITTI dataset.}
    \label{tab:ablation-cmps}
    \vspace{-0.6em}
\begin{adjustbox}{width=0.84\columnwidth}    
     \begin{tabular}{cccc}
        \toprule
        \multirow{1}{*}{\qquad Method} 
        & RE($^{\circ}$) $\downarrow$ & TE(cm) $\downarrow$ & RR(\%) $\uparrow$ \\ 
        \midrule
        1. Surfel w/o uncertainty weight & $1.85$ & $6.73$ & $91.64$ \\
        \hline
        2. Point cloud + vanilla E2PN & $0.84$ & $8.15$ & $94.53$ \\
        \hline
        3. Point Cloud + 3D CNN  & $12.08$ & $14.75$ & $56.47$\\%
        \hline
        4. Point Cloud + PointNet++ \cite{qi2017pointnet++} & $8.73$ & $10.52$ & $68.91$ \\
        \hline
        
        5. W/o attention module & $8.62$ & $18.24$ & $56.31$ \\
        \hline
        \hline
        6. $\mathcal{L}_1$ loss only & $1.26$ & $7.85$ & $90.04$\\ 
        \hline
        7. $\mathcal{L}_2$ loss only & $2.49$ & $8.93$ & $82.80$\\ 
        \hline
        \hline
        8. Full model + Huber Loss & $\mathbf{0.25}$ & $\mathbf{7.03}$ & $\mathbf{99.85}$ \\
        \bottomrule
    \end{tabular}
\end{adjustbox}    
\vspace{-1.2em}
\end{table}
\begin{table}[!thbp]
\vspace{-0.0em}
\centering
\caption{\small{Robustness analysis of average errors on KITTI.}}
\label{tab:robustness-analysis}
    \vspace{-0.6em}
\begin{small}
\begin{adjustbox}{width=0.90\columnwidth}
\begin{tabular}{c c c c c c}
\toprule
 & $5^{\circ}, 10cm$ & $25^{\circ}, 50cm$ & $50^{\circ}, 100cm$ & $75^{\circ}, 150cm$ & $100^{\circ}, 200cm$\\%
\hline
RE($^{\circ}$) $\downarrow$ & 0.18 & 0.71 & 2.37 & 3.12 &  4.60\\ 
\hline
TE(cm) $\downarrow$ & 0.51 & 1.75 & 2.86 & 6.73 & 12.59 \\
\hline
RR(\%) $\uparrow$ & 99.90 & 99.06 & 96.51 & 88.37 & 84.25 \\
\hline
\end{tabular}
\end{adjustbox}
\end{small}
\end{table}

\begin{table}[!thbp]
\vspace{-1.em}
\centering
\caption{\small{Ablation study of model complexity on KITTI.}}
\label{tab:model-complexity}
    \vspace{-0.6em}
\begin{adjustbox}{width=0.88\columnwidth}
\begin{tabular}{c c c c c c c}
\hline
 & DGR & PointDSC & Spinnet & RoReg & MAC+Trans & Ours\\%
\hline
Latency(s) $\downarrow$ & 1.26 & $\mathbf{0.08}$ & 2.84 & 22226 & 0.31 & 0.09 \\ 
\hline
Params(Mb) $\downarrow$ & 1.94 & 1.07 & 3.16 & 83.25 & 28.64 & $\mathbf{0.98}$ \\ 
\hline
\end{tabular}
\end{adjustbox}
\end{table}
Our model has a total inference time of 0.090 seconds (latency) per source-target point scan (2048 points). Additionally, we present a more detailed runtime and module size analysis of the entire model in \cref{tab:model_size}. The primary inference time is attributed to the E2PN encoder. 

\begin{table}[!th]
\vspace{-0.2em}
\centering
\caption{Summary of each module's parameter count and inference time.}
\begin{small}
\begin{adjustbox}{width=0.85\columnwidth}    
\begin{tabular}{ c c c c c}
  \toprule
   & E2PN encoder & Attention &  Decoder & Total\\ %
  \hline
  Inference time(s) & 0.057  & 0.026  & 0.007 & 0.090\\ 
  \hline
  Params(MB) & 0.681  & 0.215 & 0.084 & 0.980\\ %
  \bottomrule
\end{tabular}
\end{adjustbox} 
\end{small}
\label{tab:model_size}
\vspace{-1.0em}
\end{table}
\section{Conclusion}
We present a surfel-based $\mathbf{SE(3)}$-equivariant network designed for robust 3D registration, leveraging surfel initialization from raw depth maps or LiDAR point clouds. Our approach integrates a shared E2PN encoder, a cross-attention module, and an MLP-based decoder. Experimental results on two datasets demonstrate the strong performance of the model, highlighting its potential for real-world applications in 3D reconstruction, mapping, and augmented reality. Future work could extend this framework to more complex scenarios, such as large-scale scene reconstruction and robust registration under extreme occlusions or sparse views. Exploring dynamic scene understanding is another promising direction.

\small
\bibliographystyle{IEEEtranN}
\bibliography{references}

\end{document}